\documentclass{article} 
\usepackage{plain,times} 
\usepackage{easyReview}
\usepackage{adjustbox}
\usepackage{amsmath}
\usepackage{amssymb}
\usepackage{bbold}
\usepackage{tikz} 
\usepackage{subcaption}
\usepackage{graphicx}
\usepackage{wrapfig}

\usepackage{amsmath,amsfonts,bm}

\def\Secref#1{Section~\ref{#1}}

\def\eqref#1{equation~\ref{#1}}

\def\1{\bm{1}}

\DeclareMathAlphabet{\mathsfit}{\encodingdefault}{\sfdefault}{m}{sl}
\SetMathAlphabet{\mathsfit}{bold}{\encodingdefault}{\sfdefault}{bx}{n}

\DeclareMathOperator*{\argmin}{arg\,min}

\usepackage{hyperref}
\hypersetup{
    colorlinks=true,
    linkcolor=blue,
    urlcolor=blue,
    citecolor=blue,
    pdftitle={Overleaf Example},
    pdfpagemode=FullScreen,
    }

\title{Stable Update of Regression Trees}
\preprintcopy

\author{Morten Blørstad \thanks{Department of Informatics, University of Bergen, 5006 Bergen, Norway.}   
\And 
Berent Å. S. Lunde \thanks{Equinor, 5254 Sandsli, Norway.}  
\And 
Nello Blaser \footnotemark[1] 
}

\begin{document}

\maketitle

\begin{abstract}
Updating machine learning models with new information usually improves their predictive performance, yet, in many applications, it is also desirable to avoid changing the model predictions too much. This property is called stability.
In most cases when stability matters, so does explainability. We therefore focus on the stability of an inherently explainable machine learning method, namely regression trees. 
We aim to use the notion of \textit{empirical stability} and design algorithms for updating regression trees that provide a way to balance between predictability and empirical stability.
To achieve this, we propose a regularization method, where data points are weighted based on the uncertainty in the initial model. The balance between predictability and empirical stability can be adjusted through hyperparameters. 
This regularization method is evaluated in terms of loss and stability and assessed on a broad range of data characteristics. The results show that the proposed update method improves stability while achieving similar or better predictive performance. 
This shows that it is possible to achieve both predictive and stable results when updating regression trees. 
  \footnote{A preliminary version of results in this paper appeared in the Master thesis of the first author \citep{Blrstad2023}.}
\end{abstract}

 \section{Introduction}\label{intro}

Training an initial machine learning (ML) model is just the start of the ML life cycle. The model can be improved as more data becomes available, leading to an iterative process of continuous improvement. Thus, in industry settings, ML models are frequently updated to incorporate new information. The main objective of such a process has traditionally been to improve the model's predictive capability. However, many industries, for instance in finance or medicine, have additional requirements such as interpretability and stability. Interpretability is important due to "algorithmic accountability" (\cite{Kaminski2021}), whereas stability is crucial to ensure consistent and reliable predictions on new unseen data. In this manuscript, we focus on the stability aspect.

Achieving stability involves minimizing model variance as stability is connected to generalization and inversely related to variance (\cite{Bousquet2002}). There have been numerous studies on the stability of learning algorithms (see, e.g., \cite{Devroye1979}; \cite{Kearns1999}; \cite{Bousquet2002}) focusing on theoretical bounds of their generalization error. 
Some previous studies have also presented empirical stability measures for both classification and regression tasks, which can be divided into structural and semantic stability (\cite{Turney1995}; \cite{Briand2009}; \cite{Yang2002}; \cite{Lim2013}; \cite{Philipp2017}). We focus on semantic stability, (dis-)similarity of the predictions of two models, which we review in \Secref{updateStability}. While \emph{measuring} stability has been studied thoroughly, there are also some studies focusing on improving it for classifiers. The studies by \cite{Fard2016}, \cite{Goh2016}, \cite{Cotter2018} and \cite{Liu2022} address the issue of unnecessary prediction change when training consecutive classifiers. \cite{Fard2016} propose a Markov Chain Monte Carlo stabilization method, whereas \cite{Goh2016}, \cite{Cotter2018} introduces different optimization constraints when training classifiers. \cite{Liu2022} improve model stability of deep-learning classifiers in the context of NLP systems. In this article, we aim to improve the empirical stability of regression trees.

Regression trees are appealing due to their inherently interpretable structure, and their ability to discover interaction effects and select features. However, due to regression trees' high variance, small changes in data can often lead to very different splits, which can cause inconsistent predictions, i.e., instability (\cite{Breiman1996}, \cite{Friedman2001}). This issue of instability in tree models has been addressed in previous research for decision trees. For instance, \cite{Last2002} showed using Info-Fuzzy Networks that semantically stable decision trees could be learned for classification tasks while maintaining a reasonable level of accuracy. Additionally, \cite{Dannegger2000} proposed a bootstrap-based node-level stabilizing approach for decision trees. This method was later improved by \cite{Briand2009}, demonstrating improved structural stability compared to the classical decision tree algorithm (CART). To the best of our knowledge, no work focuses on improving the empirical semantic stability of regression trees in the context of continuous updates.  



\textbf{Contributions}
Our contributions to stabilizing regression trees are as follows. 

\begin{enumerate}
    \item We incorporate stability measures as a regularization term in the loss function, providing a way of ensuring a level of stability by tuning regularization strength.  Our novel approach involves dynamically weighting the regularization for each data point based on the model's uncertainty. 
    \item We evaluate our approach on several different datasets, showing that this technique in combination with a base stability regularization achieves a better balance of the predictability-stability trade-off.  
    \item Our python-package for stable regression tree updates is available at \url{https://github.com/MortenBlorstad/StableTreeUpdates}. 
\end{enumerate}



The remainder of the paper unfolds as follows: \Secref{updateStability} introduces empirical stability and \Secref{regressionTree} introduces regression trees. \Secref{stableRegressionTreeUpdate} describes the proposed method for stable updates of regression trees. \Secref{experiments} provides the experimental setup and results on various datasets. Finally, in \Secref{discussion} we discuss the results and conclude.

 \section{Empirical Stability}\label{updateStability}



Consider the regression task of learning a function (or model) $f$ that maps the input vector $\mathbf{x}\in \mathbb{R}^m$ to the corresponding response variable $y\in \mathbb{R}$. Learning such a function involves approximating the optimal function $f^*$ from the set of the candidate function $\mathcal{F}$ by minimising the empirical loss given a \textit{loss function} $\mathcal{L}(\cdot,\cdot)$, over a training dataset $\mathcal{D}=\{\mathbf{x}_i,y_i\}_{i=1}^n$

\begin{equation*}
    f^* \approx \hat{f} = \argmin_{f\in \mathcal{F}} \frac{1}{n}\sum_{i=1}^n[\mathcal{L}(y_i,f(\mathbf{x}_i))].
\end{equation*}

In this context, stability is defined as how much a model’s predictions change due to changes in the training data, such as the removal or addition of a data point. It reflects a data point’s impact on the learning process. Most studies focus on theoretical stability guarantees, but for many ML algorithms, no such theoretical guarantees exist. We instead take an \textit{empirical approach} and consider empirical stability measures that can be assessed using test data. Similar to how we minimize empirical loss to improve predictability, we introduce an empirical instability measure that we minimize to improve stability.

Consider a trained model $f_0$ at time $t = 0$ with the training data $\mathcal{D}_0$. At $t=1$ more data, $\mathcal{D}_{\Delta t}$ becomes available. The model is updated using all available data $\mathcal{D}_{1} = \mathcal{D}_0 \cup \mathcal{D}_{\Delta t}$ to obtain the updated model $f_{1}$. The instability of the update is given by the \textit{instability function} $\mathcal{S}(\cdot,\cdot)$,   
\begin{equation*}
   \text{empirical instability} = \frac{1}{n}\sum_{i=1}^n[\mathcal{S}(f_0(\mathbf{x}_i),f_1(\mathbf{x}_i))],
\end{equation*}
where $\mathcal{S}(\cdot,\cdot)$ measures the discrepancy between the predictions of $f_0$ and $f_1$ over a test dateset. In other words, we are considering the mean pointwise empirical instability. The choice of $\mathcal{S}(\cdot,\cdot)$, similar to the choice of $\mathcal{L}(\cdot,\cdot)$, depends on your assumptions on the error distribution. Assuming Gaussian and Laplace distributions lead to the squared error \(\mathcal{S}(a, b) = (a-b)^2\) and absolute error \(\mathcal{S}(a, b) = |a-b|\), respectively. 
Another empirical instability measure that can be written as a mean pointwise empirical instability is the negative coverage probability \(\mathcal{S}(a,b,k) = - \mathbb{1}_{|a-b| \leq k}\) (\cite{Yang2002}).

 \section{Learning Regression Trees}\label{regressionTree}

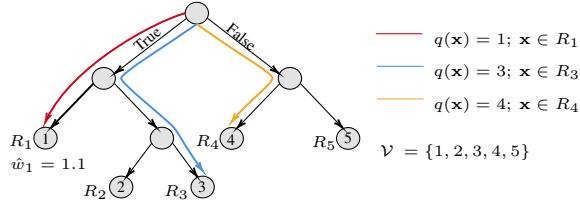
\begin{figure}[!htb]
    \centering
    
\begin{tikzpicture}[x=0.75pt,y=0.75pt,yscale=-1,xscale=1]

\draw  [fill={rgb, 255:red, 155; green, 155; blue, 155 }  ,fill opacity=0.28 ] (293.86,74.23) .. controls (293.86,71.2) and (296.49,68.74) .. (299.73,68.74) .. controls (302.97,68.74) and (305.6,71.2) .. (305.6,74.23) .. controls (305.6,77.26) and (302.97,79.71) .. (299.73,79.71) .. controls (296.49,79.71) and (293.86,77.26) .. (293.86,74.23) -- cycle ;
\draw    (294.58,76.82) -- (260.22,103.17) ;
\draw [shift={(258.64,104.39)}, rotate = 322.51] [color={rgb, 255:red, 0; green, 0; blue, 0 }  ][line width=0.75]    (4.37,-1.32) .. controls (2.78,-0.56) and (1.32,-0.12) .. (0,0) .. controls (1.32,0.12) and (2.78,0.56) .. (4.37,1.32)   ;
\draw    (305.14,76.82) -- (339.24,103.17) ;
\draw [shift={(340.82,104.39)}, rotate = 217.7] [color={rgb, 255:red, 0; green, 0; blue, 0 }  ][line width=0.75]    (4.37,-1.32) .. controls (2.78,-0.56) and (1.32,-0.12) .. (0,0) .. controls (1.32,0.12) and (2.78,0.56) .. (4.37,1.32)   ;
\draw  [fill={rgb, 255:red, 155; green, 155; blue, 155 }  ,fill opacity=0.28 ] (246.9,107.13) .. controls (246.9,104.1) and (249.52,101.65) .. (252.77,101.65) .. controls (256.01,101.65) and (258.64,104.1) .. (258.64,107.13) .. controls (258.64,110.16) and (256.01,112.62) .. (252.77,112.62) .. controls (249.52,112.62) and (246.9,110.16) .. (246.9,107.13) -- cycle ;
\draw  [fill={rgb, 255:red, 155; green, 155; blue, 155 }  ,fill opacity=0.28 ] (340.82,107.13) .. controls (340.82,104.1) and (343.45,101.65) .. (346.69,101.65) .. controls (349.94,101.65) and (352.57,104.1) .. (352.57,107.13) .. controls (352.57,110.16) and (349.94,112.62) .. (346.69,112.62) .. controls (343.45,112.62) and (340.82,110.16) .. (340.82,107.13) -- cycle ;
\draw [line width=0.75]    (247.61,109.72) -- (227.74,130.37) ;
\draw [shift={(226.35,131.81)}, rotate = 313.91] [color={rgb, 255:red, 0; green, 0; blue, 0 }  ][line width=0.75]    (4.37,-1.32) .. controls (2.78,-0.56) and (1.32,-0.12) .. (0,0) .. controls (1.32,0.12) and (2.78,0.56) .. (4.37,1.32)   ;
\draw    (258.18,109.72) -- (277.81,130.36) ;
\draw [shift={(279.18,131.81)}, rotate = 226.44] [color={rgb, 255:red, 0; green, 0; blue, 0 }  ][line width=0.75]    (4.37,-1.32) .. controls (2.78,-0.56) and (1.32,-0.12) .. (0,0) .. controls (1.32,0.12) and (2.78,0.56) .. (4.37,1.32)   ;
\draw  [fill={rgb, 255:red, 155; green, 155; blue, 155 }  ,fill opacity=0.28 ] (217.54,137.29) .. controls (217.54,134.27) and (220.17,131.81) .. (223.41,131.81) .. controls (226.66,131.81) and (229.28,134.27) .. (229.28,137.29) .. controls (229.28,140.32) and (226.66,142.78) .. (223.41,142.78) .. controls (220.17,142.78) and (217.54,140.32) .. (217.54,137.29) -- cycle ;
\draw  [fill={rgb, 255:red, 155; green, 155; blue, 155 }  ,fill opacity=0.28 ] (276.25,137.29) .. controls (276.25,134.27) and (278.88,131.81) .. (282.12,131.81) .. controls (285.36,131.81) and (287.99,134.27) .. (287.99,137.29) .. controls (287.99,140.32) and (285.36,142.78) .. (282.12,142.78) .. controls (278.88,142.78) and (276.25,140.32) .. (276.25,137.29) -- cycle ;
\draw    (341.54,109.87) -- (321.66,130.52) ;
\draw [shift={(320.28,131.96)}, rotate = 313.91] [color={rgb, 255:red, 0; green, 0; blue, 0 }  ][line width=0.75]    (4.37,-1.32) .. controls (2.78,-0.56) and (1.32,-0.12) .. (0,0) .. controls (1.32,0.12) and (2.78,0.56) .. (4.37,1.32)   ;
\draw    (352.11,109.87) -- (371.73,130.51) ;
\draw [shift={(373.11,131.96)}, rotate = 226.44] [color={rgb, 255:red, 0; green, 0; blue, 0 }  ][line width=0.75]    (4.37,-1.32) .. controls (2.78,-0.56) and (1.32,-0.12) .. (0,0) .. controls (1.32,0.12) and (2.78,0.56) .. (4.37,1.32)   ;
\draw  [fill={rgb, 255:red, 155; green, 155; blue, 155 }  ,fill opacity=0.28 ] (311.47,137.45) .. controls (311.47,134.42) and (314.1,131.96) .. (317.34,131.96) .. controls (320.58,131.96) and (323.21,134.42) .. (323.21,137.45) .. controls (323.21,140.48) and (320.58,142.93) .. (317.34,142.93) .. controls (314.1,142.93) and (311.47,140.48) .. (311.47,137.45) -- cycle ;
\draw  [fill={rgb, 255:red, 155; green, 155; blue, 155 }  ,fill opacity=0.28 ] (370.18,137.45) .. controls (370.18,134.42) and (372.81,131.96) .. (376.05,131.96) .. controls (379.29,131.96) and (381.92,134.42) .. (381.92,137.45) .. controls (381.92,140.48) and (379.29,142.93) .. (376.05,142.93) .. controls (372.81,142.93) and (370.18,140.48) .. (370.18,137.45) -- cycle ;
\draw    (276.97,139.88) -- (265.71,154.89) ;
\draw [shift={(264.51,156.49)}, rotate = 306.88] [color={rgb, 255:red, 0; green, 0; blue, 0 }  ][line width=0.75]    (4.37,-1.32) .. controls (2.78,-0.56) and (1.32,-0.12) .. (0,0) .. controls (1.32,0.12) and (2.78,0.56) .. (4.37,1.32)   ;
\draw    (287.53,139.88) -- (298.55,154.88) ;
\draw [shift={(299.73,156.49)}, rotate = 233.7] [color={rgb, 255:red, 0; green, 0; blue, 0 }  ][line width=0.75]    (4.37,-1.32) .. controls (2.78,-0.56) and (1.32,-0.12) .. (0,0) .. controls (1.32,0.12) and (2.78,0.56) .. (4.37,1.32)   ;
\draw  [fill={rgb, 255:red, 155; green, 155; blue, 155 }  ,fill opacity=0.28 ] (255.7,161.97) .. controls (255.7,158.94) and (258.33,156.49) .. (261.57,156.49) .. controls (264.81,156.49) and (267.44,158.94) .. (267.44,161.97) .. controls (267.44,165) and (264.81,167.46) .. (261.57,167.46) .. controls (258.33,167.46) and (255.7,165) .. (255.7,161.97) -- cycle ;
\draw  [fill={rgb, 255:red, 155; green, 155; blue, 155 }  ,fill opacity=0.28 ] (296.8,161.97) .. controls (296.8,158.94) and (299.42,156.49) .. (302.67,156.49) .. controls (305.91,156.49) and (308.54,158.94) .. (308.54,161.97) .. controls (308.54,165) and (305.91,167.46) .. (302.67,167.46) .. controls (299.42,167.46) and (296.8,165) .. (296.8,161.97) -- cycle ;
\draw [color={rgb, 255:red, 208; green, 2; blue, 27 }  ,draw opacity=1 ][line width=0.75]    (293.86,74.23) .. controls (277.05,78.95) and (236.87,101.16) .. (224.16,127.45) ;
\draw [shift={(223.41,129.07)}, rotate = 293.62] [color={rgb, 255:red, 208; green, 2; blue, 27 }  ,draw opacity=1 ][line width=0.75]    (4.37,-1.32) .. controls (2.78,-0.56) and (1.32,-0.12) .. (0,0) .. controls (1.32,0.12) and (2.78,0.56) .. (4.37,1.32)   ;
\draw [color={rgb, 255:red, 74; green, 144; blue, 226 }  ,draw opacity=1 ][line width=0.75]    (299.73,79.71) .. controls (288.45,87.82) and (261.44,104.27) .. (261.57,107.13) .. controls (261.7,110) and (284.4,127.55) .. (287.99,131.81) .. controls (291.45,135.93) and (292.3,139.48) .. (301.62,152.31) ;
\draw [shift={(302.67,153.75)}, rotate = 233.67] [color={rgb, 255:red, 74; green, 144; blue, 226 }  ,draw opacity=1 ][line width=0.75]    (4.37,-1.32) .. controls (2.78,-0.56) and (1.32,-0.12) .. (0,0) .. controls (1.32,0.12) and (2.78,0.56) .. (4.37,1.32)   ;
\draw [color={rgb, 255:red, 245; green, 166; blue, 35 }  ,draw opacity=1 ][line width=0.75]    (299.73,79.71) .. controls (322.89,96.59) and (338.02,104.15) .. (337.89,107.13) .. controls (337.77,109.94) and (325.31,119.15) .. (318.56,127.49) ;
\draw [shift={(317.34,129.07)}, rotate = 305.98] [color={rgb, 255:red, 245; green, 166; blue, 35 }  ,draw opacity=1 ][line width=0.75]    (4.37,-1.32) .. controls (2.78,-0.56) and (1.32,-0.12) .. (0,0) .. controls (1.32,0.12) and (2.78,0.56) .. (4.37,1.32)   ;
\draw [color={rgb, 255:red, 245; green, 166; blue, 35 }  ,draw opacity=1 ]   (390.59,117.51) -- (414.07,117.51) ;
\draw [color={rgb, 255:red, 74; green, 144; blue, 226 }  ,draw opacity=1 ]   (390.59,101.06) -- (414.07,101.06) ;
\draw [color={rgb, 255:red, 208; green, 2; blue, 27 }  ,draw opacity=1 ]   (390,84.61) -- (413.48,84.61) ;

\draw (245.75,103.97) node [anchor=north west][inner sep=0.75pt]  [font=\tiny] [align=left] {\begin{minipage}[lt]{8.67pt}\setlength\topsep{0pt}
\begin{center}

\end{center}

\end{minipage}};
\draw (300,74.7) node  [font=\tiny] [align=left] {\begin{minipage}[lt]{8.67pt}\setlength\topsep{0pt}
\begin{center}

\end{center}

\end{minipage}};
\draw (346.69,107.53) node  [font=\tiny] [align=left] {\begin{minipage}[lt]{8.67pt}\setlength\topsep{0pt}
\begin{center}

\end{center}

\end{minipage}};
\draw (216.5,133.5) node [anchor=north west][inner sep=0.75pt]  [font=\tiny] [align=left] {\begin{minipage}[lt]{8.67pt}\setlength\topsep{0pt}
\begin{center}
1
\end{center}

\end{minipage}};
\draw (282.12,137) node  [font=\tiny] [align=left] {\begin{minipage}[lt]{8.67pt}\setlength\topsep{0pt}
\begin{center}

\end{center}

\end{minipage}};
\draw (310,134) node [anchor=north west][inner sep=0.75pt]  [font=\tiny] [align=left] {\begin{minipage}[lt]{8.67pt}\setlength\topsep{0pt}
\begin{center}
4
\end{center}

\end{minipage}};
\draw (376,137) node  [font=\tiny] [align=left] {\begin{minipage}[lt]{8.67pt}\setlength\topsep{0pt}
\begin{center}
5
\end{center}

\end{minipage}};
\draw (254.5,158) node [anchor=north west][inner sep=0.75pt]  [font=\tiny] [align=left] {\begin{minipage}[lt]{8.67pt}\setlength\topsep{0pt}
\begin{center}
2
\end{center}

\end{minipage}};
\draw (302.43,162) node  [font=\tiny] [align=left] {\begin{minipage}[lt]{8.67pt}\setlength\topsep{0pt}
\begin{center}
3
\end{center}

\end{minipage}};
\draw (391,139.05) node [anchor=north west][inner sep=0.75pt]  [font=\tiny]  {$\mathcal{V} \ =\{1, 2, 3, 4, 5\}$};
\draw (204.5, 146) node [anchor=north west][inner sep=0.75pt]  [font=\tiny]  {$\hat{w}_{1} =1.1$};
\draw (418,115.3) node [anchor=north west][inner sep=0.75pt]  [font=\tiny]  {$q(\mathbf{x}) =4;\ \mathbf{x} \in R_{4}$};
\draw (418,98.85) node [anchor=north west][inner sep=0.75pt]  [font=\tiny]  {$q(\mathbf{x}) =3;\ \mathbf{x} \in R_{3}$};
\draw (418,82.4) node [anchor=north west][inner sep=0.75pt]  [font=\tiny]  {$q(\mathbf{x}) =1;\ \mathbf{x} \in R_{1}$};
\draw (203.85,134.74) node [anchor=north west][inner sep=0.75pt]  [font=\tiny]  {$R_{1}$};
\draw (282,159.7) node [anchor=north west][inner sep=0.75pt]  [font=\tiny]  {$R_{3}$};
\draw (241,159.7) node [anchor=north west][inner sep=0.75pt]  [font=\tiny]  {$R_{2}$};
\draw (297.77,134.47) node [anchor=north west][inner sep=0.75pt]  [font=\tiny]  {$R_{4}$};
\draw (356.48,135.02) node [anchor=north west][inner sep=0.75pt]  [font=\tiny]  {$R_{5}$};
\draw (265.65,89.8) node [anchor=north west][inner sep=0.75pt]  [font=\tiny,rotate=-320] [align=left] {{\tiny True}};
\draw (317.52,76.97) node [anchor=north west][inner sep=0.75pt]  [font=\tiny,rotate=-40] [align=left] {{\tiny False}};

\end{tikzpicture}

    \caption[The Classification and Regression Tree]{An example of a Classification and Regression Tree (CART) with five leaf nodes $\mathcal{V}$. The vector $\hat{\mathbf{w}} = (\hat{w}_1,\hat{w}_{2},\hat{w}_{3},\hat{w}_4,\hat{w}_5)$ is the possible predictions the tree can make.} 
    \label{fig:CART}
\end{figure}

There are many methods of fitting Regression trees, most notably the Classification And Regression Tree (CART) algorithm by \cite{Breiman1984} (see Figure \ref{fig:CART}). The CART algorithm uses a top-down greedy approach to partition the feature space $R = \mathbb{R}^m$ into $D$ non-overlapping regions, $R_1, \ldots, R_d, \ldots, R_D$ and fit a leaf prediction, $\hat{w}_d$ to each region based on the response observations within the region. Starting from a root node \(R_{root}\) containing all the data, the algorithm seeks the splitting variable $j \in {1, \ldots, m}$ and split value $s$ that creates the half-planes $R_l(j, s) =\{x \in R_{root}|x_j \leq s\}$ and $R_r(j, s) = \{x\in R_{root}|x_j > s\}$ that maximize the loss reduction $\mathcal{R}$,
\begin{equation*}
\mathcal{R}(j,s) = \sum_{i:x_i\in R_{root}}{\mathcal{L}(y_i, \hat{w}_{root})}  - \sum_{i:x_i\in R_l(j,s)}{\mathcal{L}(y_i, \hat{w}_{l})} - \sum_{i:x_i\in R_r(j,s)}{\mathcal{L}(y_i, \hat{w}_{r})}.
\end{equation*}
 Here $R_l$ and $R_r$ are the datasets of left and right nodes after the split and $R_{root} = R_l \cup R_r$ is the dataset before the split. The terms $\hat{w}_{root}$, $\hat{w}_l$ and $\hat{w}_r$ denote the predictions of the root, left and right node. After finding the partition with the largest loss reduction, the splitting process is repeated on the resulting regions until reaching a stopping criterion. 
Given a learned tree model $f$, the feature mapping $q(\mathbf{x})$ is a function that maps inputs to their corresponding region $R$ and its prediction for input \(x\) is given as $f(\mathbf{x}) = \hat{w}_{q(\mathbf{x})}$. 

 \section{Stable Regression Tree Update}\label{stableRegressionTreeUpdate}

We propose an update method that incorporates a stability regularization term into the loss function. The key lies in the determination of the regularization strength, which we propose to be a composite of base regularization and uncertainty-weighted regularization, each adjusted through a hyperparameter. This section first describes the modification made to the CART algorithm followed by a description of our proposed update method. 


We modify the CART by using the second-order approximation of the loss reduction, 
\begin{equation*}
    \tilde{\mathcal{R}}(j,s) = \frac{1}{2}\left[\frac{(\sum_{i:\mathbf{x}_i \in R_l} g_i)^2}{\sum_{i:\mathbf{x}_i \in R_l} h_i} +\frac{(\sum_{i:\mathbf{x}_i \in R_r} g_i)^2}{\sum_{i:\mathbf{x}_i \in R_r} h_i} - \frac{(\sum_{i:\mathbf{x}_i \in R_{root}} g_i)^2}{\sum_{i:\mathbf{x}_i \in R_{root}} h_i} \right],
\end{equation*}
where $g_{i} = \partial_{\hat{y}^{k-1}}\mathcal{L}(y_i, 0)$ and $h_{i} = \partial^2_{\hat{y}^{k-1}}\mathcal{L}(y_i, 0)$.
This is equivalent to a gradient boosting iteration from a model $f(\mathbf{x})^{k-1} = \hat{y}^{k-1}=0$ predicting zero and a learning rate of one (\cite{Chen2016}).
For squared error loss, this tree will be identical to a CART tree built using exact loss reduction $\mathcal{R}$. 

The second-order approximation allows the use of the \textit{adaptive tree complexity} method by \cite{Lunde2020} which adaptively stops the splitting process of a branch based on an information criterion, alleviating the need for hyperparameter tuning to optimize the tree complexity.
Additionally, it makes it easy to incorporate additional objectives to the loss function as long as they are twice differentiable. Note that as the information criterion is rooted in the loss function, any additional objectives added to the loss are factored in when determining the tree complexity.

\subsection{Stable Loss}
In our approach to improve stability, we introduce a stable loss function that incorporates a regularization term. Similar to \textit{explicit regularization} using a penalty term in the loss function to encourage simpler solutions and address the high-variance problem (i.e., overfitting), the \textit{Stable Loss} (\texttt{SL}) adds a penalty term to encourage small change in the predictions (i.e., stability). \texttt{SL} uses the predictions of the current tree $f_{0}$ on the combined dataset $\mathcal{D}_{1}= \{\mathbf{x}^{(1)}, y^{(1)}\}$ to regularize the loss function to prevent the predictions of the updated tree $f_{1}$ from deviating too much from the predictions of $f_{0}$. In other word, when calculating loss reduction \(\tilde{\mathcal{R}}\), it replaces the loss \(\mathcal{L}\) with the stability loss 
\begin{equation}
    \mathcal{L}_{\text{SL}}\left(y_i,f_{1}(\mathbf{x}_i); f_0(\mathbf{x}_i) \right) = \\
    \underbrace{\mathcal{L}(y_i,f_{1}(\mathbf{x}_i)}_{\textit{\small loss}}) + 
    \gamma_i \underbrace{\mathcal{S}(f_0(\mathbf{x}_i),f_{1}(\mathbf{x}_i)}_{\textit{\small instability}}),
    \label{eq:stable_loss}
\end{equation}
where $\gamma_i$ determines the strength of regularization (or belief in $f_0$). The choice of $\gamma_i$ is key, and we explore two distinct approaches and a combination of these to determine its value.

\subsubsection{Constant Regularization}
The simplest approach employs a uniform regularization factor $\alpha$, such that $\gamma_i = \alpha$ for every data point. Here $\alpha$ is a hyperparameter that is tuned to balance the trade-off between predictability and stability. 

\subsubsection{Uncertainty-Weigthed Regularization (\texttt{UWR})}

Intuitively, to preserve predictability, we want to be more stable for confident regions of $f_0$ compared to uncertain ones. The \texttt{UWR} dynamically adjusts $\gamma_i$ based on each data point contribution to the model's uncertainty. Abusing the notations $\hat{w}(\mathbf{x}) = \hat{w}_{q(\mathbf{x})}$, we denote $\hat{\sigma}^2_{\hat{w}_{0}(\mathbf{x})}$ as the variance estimate of the predictions \(\hat{w}_{0}(\mathbf{x})\) under the model $f_0$. Dividing the precision $\frac{1}{\hat{\sigma}^2_{\hat{w}_{0}(\mathbf{x})}}$ by the number of data points \(n(\mathbf{x})\) contributing to this variance leads to the idea  of using
\begin{equation*}
     \phi(\mathbf{x}_i) = \frac{ c}{n(\mathbf{x}_i)\hat{\sigma}^2_{\hat{w}_0(\mathbf{x}_i)}+\epsilon}.
\end{equation*}
for scaling constant \(c>0\) and small \(\epsilon>0\) to ensure numerical stability. We introduce $\beta$, a hyperparameter akin to $\alpha$, and set $\gamma_i = \beta \phi(\mathbf{x}_i)$. This is reminiscent of relative weights according to a Gaussian prior on $\hat{w}_0(\mathbf{x}_i)$. The scaling constant $c$ is chosen as the average response variance across leaf nodes, $c = n_0^{-1}\sum_{i=1} \hat{\sigma}^2_{y_i|\mathbf{x}_i}$. Assuming the variance is $O(n(\mathbf{x})^{-1})$, this places $\beta$ and $\alpha$ on a comparable scale.


The greedy splitting in CART complicates estimating $\sigma_{\hat{w}_0(\mathbf{x}_i)}^2$.
However, the relationship $C=\mathbb{E}[h_i]\sigma_{\hat{w}_d}^2$ between estimator variance and loss-based information criteria, $C$ (see \citet{TIC1976} or \citet{Burnham2004} for an English version), aids in this regard.
Specifically, \citet[Appendix A]{Lunde2020} provides a conservative estimate for $C$ correcting for greedy split points.
We use this relationship to construct the estimator for the prediction variance in node $d$:
\begin{align*}
        \hat{\sigma}_{\hat{w}_d}^2 = \frac{\sum_{i\in I_d} (g_i+h_i\hat{w}_d)^2}{(\sum_{i\in I_d} h_i)^2}\cdot\frac{1+\mathbb{E}[\max_j B_{j,root}]}{2}.
    \end{align*}
Here $B_{j,root} = \max_\pi S_j(\pi) \geq 1$ represents the maximum of a specific Cox-Ingersoll-Ross process $S_j(\pi)$ (\cite{CIR1985}), 
with $\pi$ indicating the data fraction passed to node $d$ from its parent with splitting variable $j$.
The initial term is the Huber Sandwich Estimator \citep{Huber1967}, suitable if the tree structure $q$ wasn't learned from $\mathcal{D}_0$.
The latter adjustment term grows with the number of possible splits.

\subsubsection{Combined Regularization}
We can also combine the constant regularization and \texttt{UWR} and define $\gamma_i = \alpha + \beta \phi(\mathbf{x}_i)$. This combination allows for a base regularization level ($\alpha$) while still adapting to the uncertainty connected to individual predictions ($\beta$). The hyperparameter pair $(\alpha, \beta)$ dictates the regularization strategy, (0,0) corresponds to the baseline model (i.e., minimizing the squared error with no regularization), whereas ($\alpha>0$, 0) and (0, $\beta>0$) corresponds to constant regularization and \texttt{UWR} strategy, respectively. The stable loss given a hyperparameter pair $(\alpha, \beta)$ results in the following estimate of a leaf prediction
 \begin{equation*}
    \hat{w}_{d} =- \frac{\sum_{i\in I_d}g_i}{\sum_{i\in I_d}h_i}.
\end{equation*}

For squared error loss, this is equivalent to
 \begin{equation*}
    \hat{w}_{d} = \frac{\sum_{i\in I_d} \left(y_i + \left(\alpha + \beta \phi(\mathbf{x}_i)\right) \hat{w}_0(\mathbf{x}_i)\right)}{\sum_{i\in I_d} \left(1+\alpha + \beta \phi(\mathbf{x}_i)\right)}.
\end{equation*}

 \section{Experiments}\label{experiments}

In this section, we evaluate the Pareto efficiency of the proposed update method with different $(\alpha,\beta)$ settings on different regression tasks. Note that because we try to achieve two objectives (loss and instability minimization) simultaneously, there is a trade-off between the two objectives. The Pareto front represents all the solutions that have the best-achieved loss for a given level of stability and vice versa. For our experiments, we use squared error loss and squared error instability. We first describe the experimental setups and then present the results. 

\subsection{Experimental Setup}

\textbf{Datasets.} We evaluate our proposed method on various benchmark regression datasets to get a comprehensive predictability-stability assessment across a broad range of data characteristics, detailed in Table \ref{table:data_overview}. 
\begin{table}[!htb]
\centering
\small 
\begin{adjustbox}{max width= 0.9\textwidth}
\begin{tabular}{l c p{7cm}} 
\hline
\textbf{Dataset} & $\mathbf{n\times m}$ & \textbf{Description} \\
\hline
\texttt{California} & $20,640 \times 8$ & \footnotesize{Aggregated housing data in the California district using one row per census block group.} \\
\texttt{Boston} & $506 \times 13$ & \footnotesize{Housing data for the suburbs of Boston.} \\
\texttt{Carseats} & $400 \times 11$ & \footnotesize{Simulated data of child car seat sales.} \\
\texttt{College} & $777 \times 18$ & \footnotesize{Statistics for a large number of US Colleges.} \\
\texttt{Hitters} & $263 \times 20$ & \footnotesize{Major League Baseball Data from the 1986 and 1987 seasons.} \\
\texttt{Wage} & $3000 \times 16$ & \footnotesize{Wage data for male workers in the Mid-Atlantic region.} \\
\hline
\end{tabular}
\end{adjustbox}
\caption{Datasets Description. The \texttt{California} dataset is adapted from the book Sparse Spatial Autoregressions, Statistics and Probability Letters (\cite{Pace1997}), whereas the other datasets are from the book ”Introduction to Statistical Learning” (ISLR) by \cite{ISLR22022}. The dimensions are after data preprocessing, i.e., one-hot encoding and removal of any missing values.}
\label{table:data_overview}
\end{table}

We perform a comprehensive grid search across 121 configurations of the hyperparameters $\alpha$ and $\beta$, where both range from 0 to 2 in increments of 0.2. For all experiments we set $\epsilon=0.01$. \textbf{Baseline.} Squared error loss, corresponds to hyperparameter setting ($\alpha = 0$, $\beta = 0$). \textbf{Constant.} Corresponds to hyperparameter setting ($\alpha >0$, $\beta = 0$). \textbf{\texttt{UWR}.} Corresponds to hyperparameter setting ($\alpha = 0$, $\beta > 0$); \textbf{Combined.} Corresponds to hyperparameter setting ($\alpha > 0$, $\beta > 0$).

\textbf{The setups.} The setups aim to reflect how a model needs to adapt when new information becomes available. \textbf{The main setup} consists of a single update iteration $t=1$. First, a model is initialized and trained on the currently available data $\mathcal{D}_0$. At $t=1$ new information is available that is added to $\mathcal{D}_0$ to obtain $\mathcal{D}_1$ and the model is updated using \texttt{SL}. 
The loss and stability are computed using repeated cross-validation. For each dataset, the dataset is divided into five folds, where each fold serves as test data once. The remaining four folds represent $\mathcal{D}_1$ and half of $\mathcal{D}_1$ represents $\mathcal{D}_0$. This process is repeated ten times and the average of the loss and stability estimates from test data is used to obtain the final loss and stability measures. 

\begin{wrapfigure}[20]{R}{0.5\textwidth}
  \vspace{-10pt}
  \centering
  \includegraphics[width=0.875\linewidth]{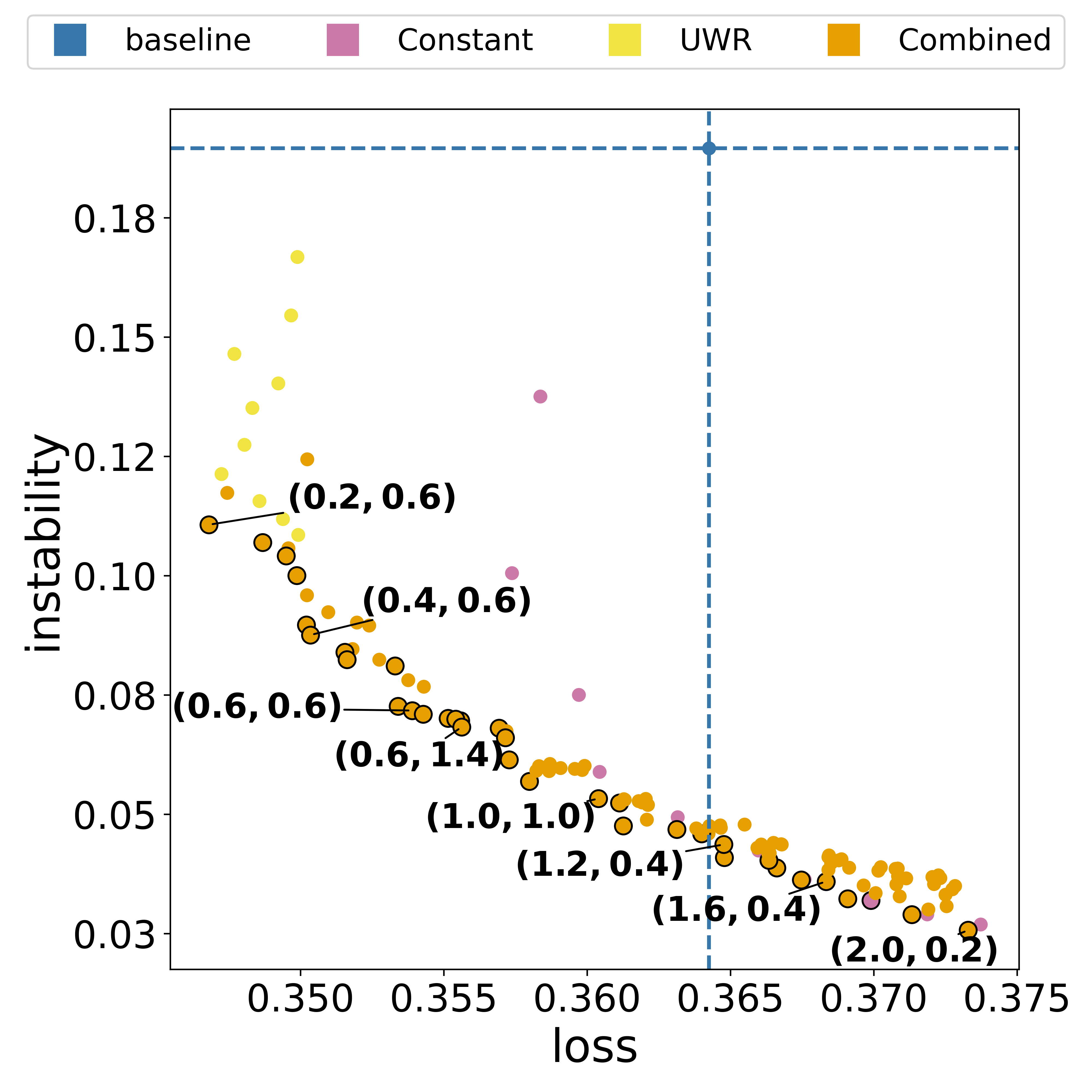}
  \vspace{-10pt}
  \caption[Pareto efficiency of the model updates over different datasets.]{The loss and stability of the model update strategies, highlighting the models that form the Pareto frontier with black borders.}\label{figParetoFrontier}
\end{wrapfigure}

\textbf{Secondary setups.} Two additional experiments are  conducted using the \texttt{California} dataset. We select five different hyperparameter configurations from the Pareto-efficient configurations identified in the main setup, each representing a different trade-off.  \textbf{The varied sample sizes setup} uses 5000 data points as test data to compute loss and stability. The dataset $\mathcal{D}_1$ is obtained by samples $n$ samples without replacement from the remaining data. Half of $\mathcal{D}_1$ represents $\mathcal{D}_0$. An initial model is trained on $\mathcal{D}_0$ and updated using $\mathcal{D}_1$. We run the experiment for different values of $n \in [1000, 5000, 10000, 15000]$ and repeat it 50 times and average loss and stability estimates on test data. \textbf{The iterative updates setup} divides the dataset into seven folds, where one fold serves as $\mathcal{D}_0$ and one serves as test data. The remaining five folds represent additional information, $\mathcal{D}_{\Delta_t}$, obtained at different update iterations. An initial model is trained on $\mathcal{D}_0$. For each update iteration a fold is added to $\mathcal{D}_0$ to obtain $\mathcal{D}_t$ and the model is updated and evaluated on the test data. The experiment is repeated 50 times and averages loss and stability estimates on test data.

\subsection{Experimental Results}

Figure \ref{figParetoFrontier} shows the results of the main experiment for the \texttt{California} dataset. The baseline is the least stable model and is not on the Pareto frontier. As we introduce regularization, more predictive and/or stable solutions can be achieved by tuning $\alpha$ and $\beta$. The constant approach shows a consistent reduction in instability as $\alpha$ increases and up to a certain $\alpha$ value, it also reduces loss. The \texttt{UWR} approach results in more stable models than the baseline with lower loss compared to both the baseline and constant approach. The combination of the constant and the \texttt{UWR} approach leads to better solutions than the pure constant and \texttt{UWR} solutions, showcased by the Pareto frontier. 

\begin{figure*}[!h]
  \centering
  \includegraphics[width=0.9\linewidth]{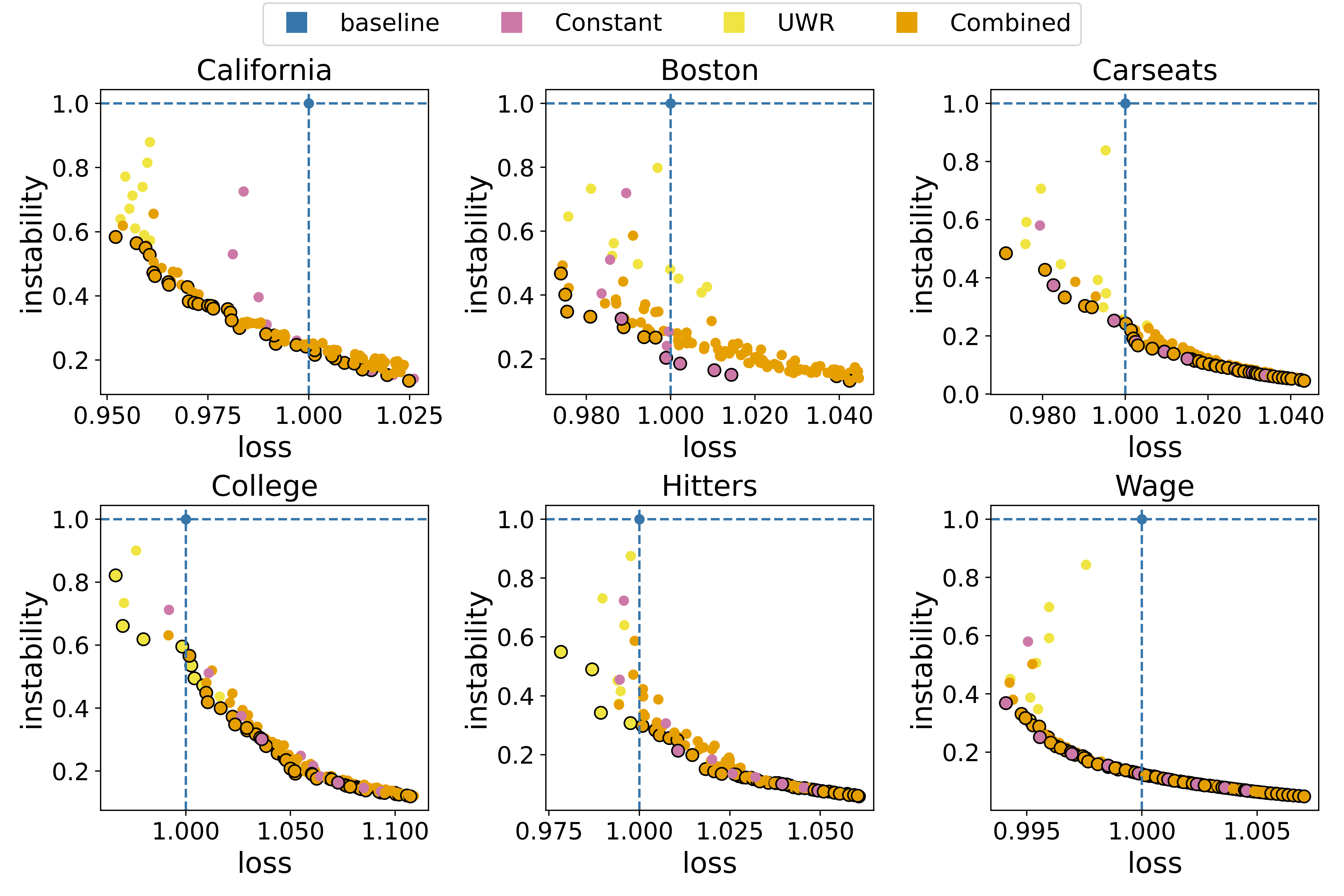}
  \caption[Pareto efficiency of the model updates over different datasets.]{The figure plots the loss and stability of the model update strategies and highlights the models that form the Pareto frontier with black borders. The loss and stability values are relative to the baseline.}\label{figParetoFrontierAll}
\end{figure*}

Figure \ref{figParetoFrontierAll} shows and compares the results of the main experiment for all six datasets. Notably, the baseline does not appear on the Pareto frontier for any of the datasets, with all ($\alpha$, $\beta$) configurations being more stable. 
Furthermore, many configurations also achieve lower loss than the baseline. Comparing the constant and \texttt{UWR} approaches, the constant approach tends to lead to more stable solutions, whereas the \texttt{UWR} approach tends to achieve lower loss, although there is an exception for \texttt{Wage} datasets. The combined approach frequently finds Pareto-efficient solutions that are unattainable by either the constant or \texttt{UWR} approaches in isolation. 

\begin{figure}[!h]
  \centering
  \begin{subfigure}[t]{0.48\textwidth}
  \centering
    \includegraphics[width=0.8\linewidth,keepaspectratio]{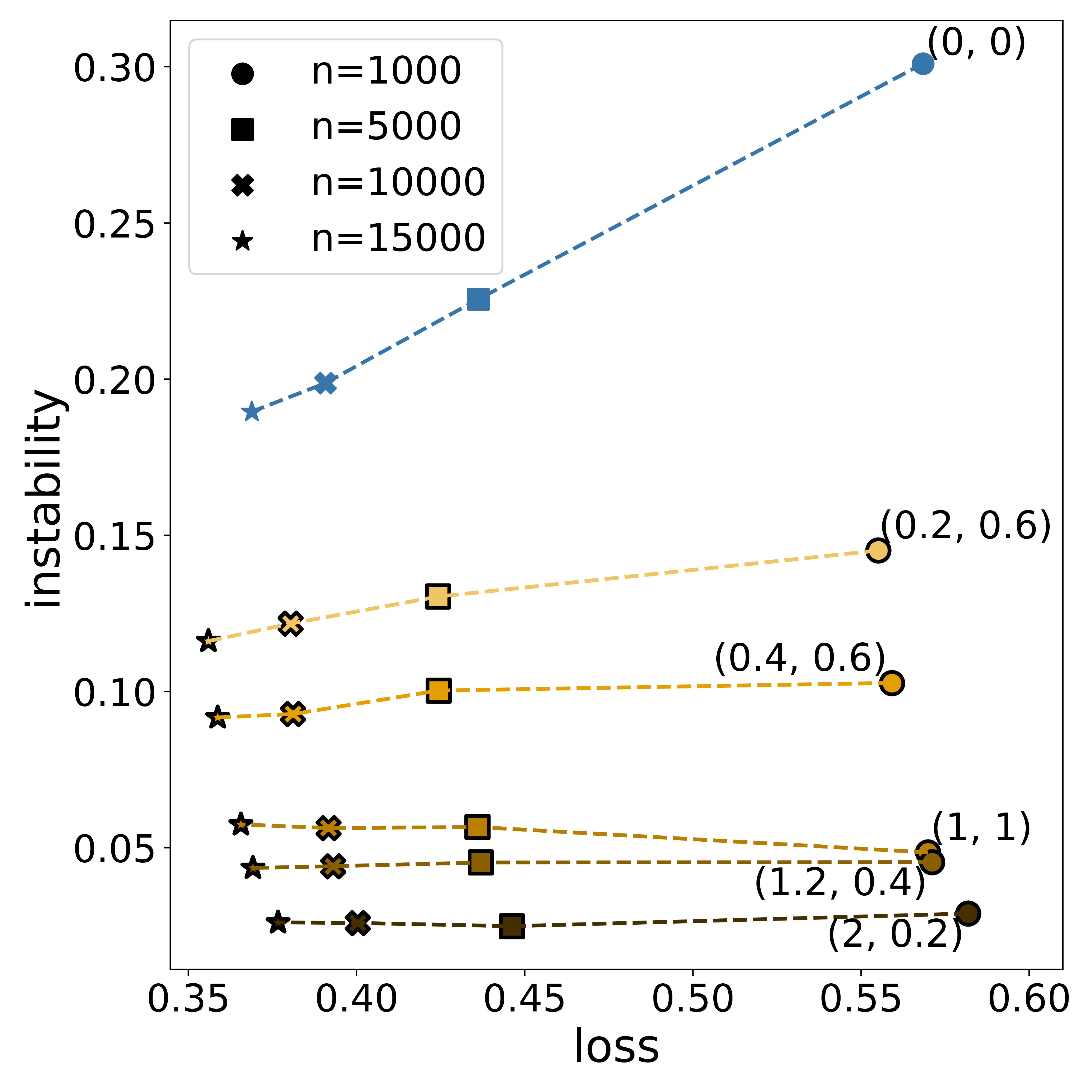}
    \caption[Comparing model updates with different sample sizes using the California Housing dataset.]{
    The varied sample sizes setup. }
    \label{figIncreasingSampleSize}
  \end{subfigure}
  \hfill
  \begin{subfigure}[t]{0.48\textwidth}
  \centering
    \includegraphics[width=0.8\linewidth,keepaspectratio]{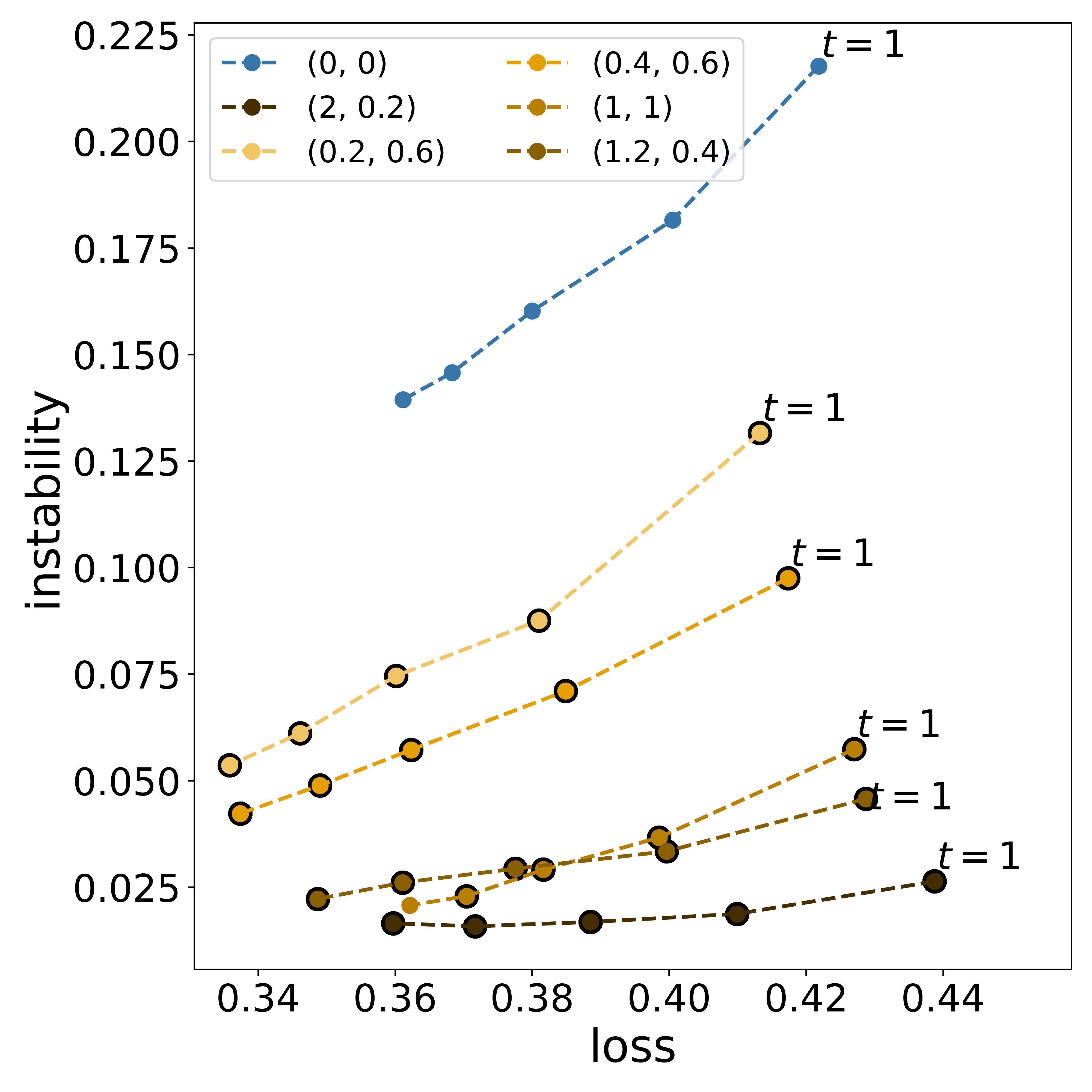}
    \caption[Tracking model updates over time using the California Housing dataset]{The iterative updates setup.}
    \label{figUpdatesOvertime}
  \end{subfigure}
  \caption{The loss-instability trade-off of model updates with different hyperparameter configurations $(\alpha,\beta)$ using the California Housing dataset, highlighting Pareto efficient solutions with black borders: (a) Effect of sample size on model updates. (b) Evolution of the loss-instability trade-off over five update iterations, where each point indicates a model's performance at a specific iteration $t$.}
  \label{figSecondaryExperiments}
\end{figure}

The result of the varied sample size experiment (Figure \ref{figIncreasingSampleSize}) shows that loss and stability decrease with larger sample sizes for baseline. For the five selected ($\alpha$, $\beta$) configurations, loss decreases while instability remains around the same level. Among the six models, all five configurations are always on the Pareto frontier for all sample sizes. The configurations are consistently more stable than the baseline at corresponding sample sizes. The configurations (0.2, 0.6) and (0.4, 0.6) also consistently achieve lower losses than the baseline.



Figure \ref{figUpdatesOvertime} presents the result of five selected ($\alpha$, $\beta$) configurations over five update iterations. Compared to the baseline, each configuration achieves improved stability at all update iterations. At the first update iteration, the configurations (0.2, 0.6) and (0.4, 0.6) achieve a lower loss than the baseline and at the second update iteration, also configurations (1, 1) and (1.2, 0.4) achieve a lower loss. The next iteration, (1, 1) becomes more stable than (1.2, 0.4) but with a higher loss than the baseline. By the final update iteration, all configurations except (1, 1) result in a more predictive solution than the baseline. At the first iteration, all configurations are Pareto-efficient solutions, where the configurations (0.2, 0.6), (0.4, 0.6), (1.2, 0.4) and (2, 0.2) are Pareto-efficient models across all update iterations. Note that in practice one would not stick to one configuration across all updates, but rather do a hyperparameter search for each update iteration.

 \section{Discussion}\label{discussion}

The objective of this paper was to provide stable updates for regression trees that better balance the predictability-stability trade-off. We propose a regularization method with two tunable hyperparameters. Through a series of experiments conducted across various datasets and conditions, we demonstrated the limitations of the baseline approach. Compared to our method, the baseline consistently fails to be on the Pareto frontier, suggesting the necessity of regularization to achieve a better trade-off. As we explore the regularization landscape by adjusting $\alpha$ and $\beta$, we observe a diverse set of Pareto efficient solutions, each representing a unique balance between predictability and stability.

The main experiment on the \texttt{California} dataset revealed that the constant approach tends to be more stability-focused as stability is equally important for each data point. Incrementally increasing the constant regularization factor $\alpha$ consistently improves model stability and up to a certain value of $\alpha$ also improves predictability. Conversely, the \texttt{UWR} approach tends to be more predictability-focused as stability for predictions in uncertain regions of $f_0$ receives less weight. The fact that \texttt{UWR} often improves predictability demonstrates the benefit of utilizing prediction uncertainty for regularization. Combining the two approaches leads us to a spectrum of Pareto-efficient solutions, where we can strategically choose ($\alpha$, $\beta$) configurations based on the desired predictability-stability trade-off. These findings showcase that both regularization approaches are relevant. 
Furthermore, these results also extend to the other datasets, demonstrating the generalizability of our approach. Across all datasets, the baseline is never one of the Pareto-efficient models. All ($\alpha$, $\beta$) configurations led to more stable models with many configurations also achieving lower loss. 

Our secondary experiments showed that our regularization method gives consistent results regardless of sample size and that the benefits of our method extend over multiple updates. When updating multiple times, our method continues to outperform the baseline in terms of predictability and/or stability. The models with lower $\alpha$ and $\beta$ values are the most predictive with steady improvements in stability, while the models with large $\alpha$ and small $\beta$ are the most stable and remain at the same stability level across updates, with steady improvements in predictability. Having large $\alpha$ and $\beta$ values seems to quickly become too stability-focused, which comes at the cost of predictability. 

A good choice of hyperparameters depends on the user's preference. Conservative users might prefer stability and therefore opt for larger $\alpha$ and/or $\beta$ values, while users primarily concerned with performance would rather opt for lower values. In practice, users should specify a stability threshold before model selection.


A notable constraint of our approach is the requirement for batch training, needing all available data to update the model. This is necessary as the CART algorithm, on which our updating strategy relies, is a batch algorithm. Future work could explore ways to remove the reliance on previous data to make the method fit in an incremental learning setting. 

The focus of our study was on regression tasks using squared error loss. It would be interesting to apply stable updates of tree models to other tasks, such as classification or survival analysis and use other types of loss and instability functions, such as log loss and Poisson loss. This would also expand the potential for industry applications.

While we show that our method improves the empirical stability of regression trees in terms of semantic stability, we do not consider structural stability in this paper. Structural stability is more directly related to explainability and therefore of substantial interest. Investigating how our method influences the structural stability of trees is a valuable direction for future research.

Our results show that iterative updates improve predictability. This leads us to the following untested hypothesis: Learning a regression tree by starting with a subset of the data and iteratively incorporating more data using our update method improves upon the CART algorithm. This raises the questions of how large the subset should be and how many updates to perform.

Lastly, our focus has been solely on individual regression trees. An interesting extension would be to investigate how our method works in an ensemble setting like random forests. 

In conclusion, stable regression trees provide a good way of balancing the predictability-stability trade-off, highlighted by its ability to outperform the baseline approach across various datasets, data sizes and multiple update iterations.

\subsubsection*{Acknowledgments} 
NB was supported in part by the Research Council of Norway grant “Parameterized Complexity for Practical Computing (PCPC)” (NFR, no. 274526).

\bibliography{references}
\bibliographystyle{plain}


\end{document}